\documentclass[letterpaper]{article} 
\usepackage{aaai2026}  
\usepackage{times}  
\usepackage{helvet}  
\usepackage{courier}  
\usepackage[hyphens]{url}  
\usepackage{graphicx} 
\usepackage{natbib}  
\usepackage{caption} 
\usepackage{algorithm}
\usepackage{algorithmic}
\usepackage{color,array}

\usepackage{graphicx}
\usepackage{amsmath}
\usepackage{amssymb}
\usepackage{mathtools}
\usepackage{amsthm}
\usepackage{amsmath, bm}
\usepackage{newfloat}
\usepackage{listings}
\DeclareCaptionStyle{ruled}{labelfont=normalfont,labelsep=colon,strut=off} 
\floatstyle{ruled}
\newfloat{listing}{tb}{lst}{}
\floatname{listing}{Listing}
\title{On the Dataless Training of Neural Networks}

\author{
  Alvaro Velasquez\textsuperscript{\rm 1} \quad
  Susmit Jha\textsuperscript{\rm 2} \quad
  Ismail R. Alkhouri\textsuperscript{\rm 3} 
}
\affiliations{
  \textsuperscript{\rm 1} Computer Science Department - University of Colorado at Boulder \\
  \textsuperscript{\rm 2} Computer Science Laboratory - SRI International \\
  \textsuperscript{\rm 3} Electrical Engineering \& Computer Science Department - University of Michigan - Ann Arbor \\  
alvaro.velasquez@colorado.edu; susmitjha@berkeley.edu; ismailal@umich.edu 
}

\begin{document}
\maketitle

\begin{abstract}
This paper surveys studies on the use of neural networks for optimization in the training-data-free setting. Specifically, we examine the dataless application of neural network architectures in optimization by re-parameterizing problems using fully connected (or MLP), convolutional, graph, and quadratic neural networks. Although MLPs have been used to solve linear programs a few decades ago, this approach has recently gained increasing attention due to its promising results across diverse applications, including those based on combinatorial optimization, inverse problems, and partial differential equations. The motivation for this setting stems from two key (possibly over-lapping) factors: (\textit{i}) data-driven learning approaches are still underdeveloped and have yet to demonstrate strong results, as seen in combinatorial optimization, and (\textit{ii}) the availability of training data is inherently limited, such as in medical image reconstruction and other scientific applications. In this paper, we define the dataless setting and categorize it into two variants based on how a problem instance—defined by a single datum—is encoded onto the neural network: (\textit{i}) architecture-agnostic methods and (\textit{ii}) architecture-specific methods. Additionally, we discuss similarities and clarify distinctions between the dataless neural network (dNN) settings and related concepts such as zero-shot learning, one-shot learning, lifting in optimization, and over-parameterization. 
\end{abstract}


\section{Introduction}

Neural networks (NNs) \citep{rumelhart1986learning} have been central to almost all recent breakthroughs in deep learning and artificial intelligence \citep{nielsen2015neural}. Their impact is exemplified by the capabilities of modern models such as large language models (LLMs) \citep{radford2018improving, radford2019language, brown2020language} and diffusion models \citep{song2019generative, ho2020denoising, rombach2022high}. Furthermore, deep learning techniques have enabled significant advances across a wide range of applications—from protein folding \citep{jumper2021highly} to algorithm discovery systems \citep{novikov2025alphaevolve}.

While training deep models has traditionally relied on access to large, often labeled datasets, this assumption does not hold in many scientific and engineering domains where data may be limited, expensive to acquire, or unavailable altogether \citep{alkhouri2024image, liang2025ugodit, cobb2024direct}. To address this, alternative uses of neural networks have emerged in which the model is optimized with respect to (w.r.t.) a single datum representing a specific problem instance. This approach has been successfully applied in diverse contexts, including inverse problems \citep{ulyanov2018deep}, integer programming \citep{NNforLPs1995}, combinatorial optimization \citep{alkhouri2022boss}, and partial differential equations \citep{sirignano2018dgm}.

In \citep{alkhouriRelu}, such use of neural networks is formalized under the term \textbf{dataless neural networks (dNNs)}—referring to frameworks where neural networks are optimized (parameters and/or input) using \textit{only data from a single problem instance}, without requiring any external training data. This is achieved by training the dataless neural network in reference to the graph structure and constraints of the given problem instance as opposed to relying on a dataset of related problems and solutions. This paradigm stands in contrast to conventional models learning and represents a growing line of research in the training-data-free setting.

\subsection{Contributions}
In this paper, we survey dNN methods from their early inception in the 1990s to the most recent developments. We organize existing approaches based on how the problem is embedded within the network, distinguishing between (\textit{i}) architecture-agnostic methods, where the problem is encoded in the loss function applied only to the output of the network, and (\textit{ii}) architecture-specific methods, where the problem is encoded in the architecture of the network. We also discuss similarities and differences between dNNs and related paradigms such as zero-shot learning, one-shot learning, and lifting-based methods. 









\section{Definition, categorization, and related techniques}\label{sec: def and cat}

Recent neural networks mostly consist of multiple layers that are in use in deep learning such as fully connected, convolutional, and attention layers. Furthermore, other functions are used such as activation functions which include, but not limited to, relu, sigmoid, and Gumbel softmax \cite{kunc2024three}. 

Let function $f$ represents a neural network parameterized by a set of weights, $\theta$. Generally, in supervised, unsupervised, and reinforcement learning, a training set $D_{SL} = \{(x_i, y_i)\}_i, D_{UL} = \{x_i\}_i, D_{RL} = \{(s_t, a_t, s_{t+1}, r_{t+1})\}_t$ is used to optimize $\theta$ using a loss function $\mathcal{L}$. \textbf{dataless Neural Networks (dNNs)} are defined as neural networks whose loss functions do not depend on training data (i.e., $D_{dNN} = \emptyset$) and are minimized using only the problem instance as the singular piece of data. For example, given a graph optimization problem encoded as a graph $G$ and constraints $c$, the parameters and loss function of the dNN can be defined in reference to said problem by mappings $G \times c \to \theta$ and $G \times c \to \mathcal{L}$. In other words, dNNs are NN-based techniques that are not learning over training data and is therefore different from supervised, generative, unsupervised, and reinforcement learning.

Based on how the data/information from a single problem instance is embedded into the network, we categorize dNNs into two types: 

\begin{itemize}
    \item \textbf{Architecture-agnostic methods:} Here, the information from the given problem instance is encoded into only the loss function using the output of the network. In other words, the architecture of the problem is generic and does not encode any information from the single problem of interest.

    \item \textbf{Architecture-specific methods:} Here, the information from the single datum is directly included into the architecture using standard layers such as fully connected layers. 


\end{itemize}

\subsection{Distinctions from zero- and one-shot learning, and lifting}

It is worth noting that dNNs differ from zero-shot and one-shot machine learning methods. This distinction is nuanced, but critical. While neither dNNs nor zero-shot learning architectures rely on a dataset, the latter requires a pre-trained model that was trained on a dataset before the zero-shot inference techniques can be applied \cite{pourpanah2022review}. Moreover, zero-shot learning does not entail training the zero-shot architecture, whereas dNNs do train their architecture in reference to the structure of the given problem as opposed to a dataset. Similarly, one-shot learning relies on a pre-trained architecture and a single data point to fine-tune said architecture over the target data distribution \cite{song2023comprehensive}. The reliance of dNNs on the structure of the given problem definition means that we can treat the problem instance as the single piece of data used to define the architecture or loss function of the dNN. However, dNNs are no more a one-shot learning architecture than any satisfiability or integer programming solver would be when they receive a single problem formulation to solve.

\textcolor{black}{Lifting in optimization refers to reformulating an optimization problem into a higher-dimensional space, where auxiliary variables are introduced so that nonlinear or discrete constraints become representable by linear or convex relations \cite{ben2001lectures}. Classical examples include semidefinite and polynomial liftings, where quadratic or higher-order interactions are expressed via matrix or tensor variables, enabling tractable convex relaxations. The intuition is that optimization in a lifted space can expose latent structure that is hidden in the original formulation, often allowing tighter relaxations or smoother search landscapes. Dataless neural networks share a conceptual similarity with lifting in that both introduce implicit representations to reformulate a problem instance. In dNNs, however, the lifting is implemented through the neural network’s parameterization—its weights and activations—rather than by adding explicit auxiliary variables. This neural parameterization defines a continuous manifold of candidate solutions that is optimized with respect to the given problem instance, while classical lifting defines an algebraic embedding typically governed by constraints.}

\section{Dataless NN-based methods}

In this section, we provide an overview of dNN methods based on the problems they are solving which ranges from broad settings such as Linear Programming to a narrower set of problems such as graph-based NP-hard problems. 


\subsection{Linear programming}

In linear programming (LP), the single problem instance is the given data (known vectors and matrices) in the objective and constraints. 

The 1995 paper \citep{NNforLPs1995} explores how different neural networks can be used to approximate solutions to some LP problems. The authors design networks where the problem constraints and objective are encoded into the network’s energy function and architecture, so that the network dynamics converge toward feasible and optimal points. Since the problem is explicitly embedded into the neural network, this approach aligns with architecture-specific dNN. We note that the methods in the study did not explicitly use the now-common layers; instead, they used integrators and other functions such as the signum function \cite{oppenheim1997signals}.

Empirically, the paper compares several network models on different LP examples, reporting that while convergence to valid solutions is possible, the methods differ in speed and accuracy. Overall, the results show that neural formulations of LP are feasible but come with trade-offs relative to classical LP techniques such as the simplex method \citep{dantzig1951maximization}.


\subsection{Quadratic programming}

The 1999 paper \citep{798504} presents a neural network formulation/method for continuous quadratic optimization problems with linear constraints. The data from the quadratic program (i.e., objective function and the set of constraints) represents the single datum in dNNs. The proposed approach constructs a neural network whose energy function encodes both the quadratic objective and the linear constraints, so that the network’s dynamic evolution seeks stable states corresponding to feasible solutions. Because the optimization problem is directly embedded into the architecture, this method falls under the architecture-specific dNN category. Similar to the LP dNN \citep{NNforLPs1995} in the previous subsection, this method used integrators and the sigmoid function \citep{goodfellow2016deep}.


Empirically, the authors test the approach on two examples of quadratic optimization instances, showing that the neural networks can recover reasonable solutions with acceptable accuracy, though the convergence properties and efficiency vary depending on the network design and parameter settings. Overall, the study demonstrates that the proposed method can serve as an alternative solver for quadratic optimization, while also highlighting practical limitations compared to established optimization algorithms.


\subsection{NP-hard graph combinatorial problems}

In this subsection, we provide an overview of two dNN approches for NP-hard graph problems. Here, the single datum is the graph of interest. 

The 2021 paper \citep{schuetz2022combinatorial} introduces a dNN approach for solving the Max-Cut (MaxCut) and Maximum Independent Set (MIS) problems. The method encodes the problem graph directly into a graph neural network (GNN) through the message passing technique \citep{gilmer2017neural}, and the optimization objective is represented as a physics-inspired (PI) Hamiltonian (and hence the term PI-GNN). Because the graph structure is explicitly encoded in the network architecture, this method belongs to the architecture-specific dNN category. The method is motivated by the analogy between energy minimization in statistical physics and objective minimization in combinatorial optimization. Empirically, using some graph datasets, the authors show that their approach can scale to large problem sizes and achieves solution qualities competitive with classical heuristics. Overall, the work highlights both the potential and the limitations of encoding problem structure directly into neural architectures for the training-data-free optimization. 

A series of other PI-GNN papers followed the work in \citep{schuetz2022combinatorial} such as the recent work in \citep{ichikawa2024controlling} where the authors proposed to use convex annealing in the loss function to improve exploration. 

The 2022 paper \citep{alkhouriRelu} proposes a method that frames the MIS problem as a differentiable optimization task. The approach uses a relu-based two-layer neural network architecture that defines the MIS constraints and objective through a continuous relaxation, with optimization carried out directly on a subset of the network parameters for each problem instance. Since the graph structure is embedded in the architecture, this method belongs to the architecture-specific dNN category. The intuition is that by making the MIS objective differentiable, one can leverage adaptive gradient-based optimization such as ADAM \citep{adam2014method} to search efficiently over relaxed solution spaces. Theoretically, the authors proved correctness of the proposed function. Empirically, the paper reports competitive independent sets on some benchmark graphs. 

The relu-based dNN approach in \citep{alkhouri2024dataless} was later extended to several other NP-hard problems in \citep{relu_dnn_mani_several} although these methods were not empirically verified. 



\subsection{NP-hard satisfiability problems}

The 2023 paper \citep{hosny2024torchmsat} presents a neural formulation for approximating solutions to the maximum satisfiability problem (MAX-SAT). The method models the Boolean satisfiability objective into a differentiable relaxation and optimizes it using gradient-based updates within PyTorch, making it compatible with GPU acceleration. Since the problem structure is expressed through the loss and encoded into the network itself, the approach falls into the architecture-specific dNN category. The intuition is that by relaxing Boolean variables into continuous values and leveraging parallel optimization on GPUs, one can explore large problem instances efficiently while still maintaining a connection to the discrete SAT objective. Empirically, the paper demonstrates that torchmSAT can handle sizable formulas and produce approximate solutions with reasonable clause satisfaction rates, though the outcomes depend on the rounding step used to recover binary assignments. 

\textcolor{black}{Another paper is DiffSAT \cite{zhang2024diffsat}, where the authors proposed a differential MaxSAT layer for SAT solving. DiffSAT is an approach that differentiates the discrete SAT variables and
searches for satisfying assignments through the forward and backward propagation of an NN layer, combined with  semidefinite approximation initialization. Their empirical results demonstrate that DiffSAT exhibits superior performance compared to existing learning-based and data-intensive SAT solvers, and can be generalized to
solve large-scale SAT problems.}


\subsection{Inverse imaging problems}

In the past decade, there have been three approaches for solving inverse imaging problems (IIPs): Deep Image Prior (DIP) methods \citep{alkhouri2025understanding}, Prior-free Implicit Neural Representation (INR) \citep{sitzmann2020implicit} methods, and zero-shot self-supervised methods \citep{yamanzero}. Here, the single datum of dNNs corresponds to the degraded image and the forward operator.

The 2018 paper \citep{ulyanov2018deep} introduced DIP, an approach where an untrained convolutional neural network is directly fitted to a single degraded image in order to perform image restoration tasks (IIP with natural images) such as denoising, inpainting, and super-resolution. The method relies on the inductive bias of the convolutional architecture itself: by optimizing the randomly initialized network to reconstruct the given degraded image, the network naturally captures low-frequency features before eventually fitting to noise or some vector in the null space of the forward operator (as was shown in the recent paper \citep{shijun_self_guided}). 
Because the IIP instance is only encoded in the loss function at the output of the network, this approach is best categorized as an architecture-agnostic dNN. Empirically, the paper shows that Deep Image Prior achieves restoration results that outperform a few baselines like bicubic interpolation, though it remains slower and vulnerable to overfitting. DIP was later extended to more modalities/tasks such as Magnetic Resonance Imaging (MRI) and Computed tomography (CT) \citep{alkhouri2024image}. Several more works followed \citep{ulyanov2018deep}, trying to address the overfitting as recently surveyed in the tutorial paper of \citep{alkhouri2025understanding}. 

A well-known approach for prior-free INR is the study in \citep{sitzmann2020implicit} (from 2020) where the authors proposed the use of sinusoidal representation networks (SIRENs), a type of fully connected networks that uses sine activations to model continuous signals. The method parameterizes signals such as images, audio, and 3D shapes directly as functions over coordinates, with periodic activations allowing the networks to capture fine detail and represent derivatives accurately. Since the problem information (i.e., pixels and their values) only determine the size of the input embedding, SIREN is best categorized as an architecture-agnostic dNN. The intuition behind the approach is that sine activations, unlike ReLU or tanh, naturally propagate high-frequency information and yield well-behaved derivatives, making them effective for representing complex signals and solving IIPs. Empirical results show that SIRENs outperform ReLU-based implicit networks. The work has limitations in terms of scalability and generalization fidelity. Therefore, many works followed SIREN. We refer the readers to the recent comprehensive study in \citep{kim2025grids} that evaluated several INRs methods using various metrics and tasks. 

The 2022 paper \citep{yamanzero} proposed a self-supervised method for accelerated MRI reconstruction that uses only the undersampled k-space data from the scan of a single subject, partitioning the available measurements into disjoint subsets: two for enforcing data-consistency and defining the training loss, and one for self-validation to determine early stopping. Because the problem information is entirely encoded in the loss (defined by the measurement model), the method fits within the architecture-agnostic dNN category. The intuition is that enforcing consistency with subsets of the same measurement can drive learning without ground-truth images, while the validation set prevents overfitting to noise or undersampling artifacts. Empirically, the authors demonstrate that this zero-shot approach achieves reconstructions of competitive quality compared to supervised and database-driven self-supervised methods. The study highlights both the potential of subject-specific, training-data-free reconstructions and the trade-offs in terms of computational cost during test time.

\subsection{Partial differential equations}

The 2017 paper in \citep{yu2018deep} introduces the Deep Ritz Method, which reformulates variational problems arising from Partial differential equations (PDEs) in terms of neural network–based trial functions. In this approach, a deep residual network (with fully connected layers and cubic relu activation functions) parameterizes the candidate solution, and the variational functional is approximated via stochastic quadrature, with optimization carried out using stochastic gradient descent (SGD). Because the method encodes the problem entirely in the loss, it belongs to the architecture-agnostic dNN category, where the network is a generic approximator rather than incorporating PDE-specific architectural design. The intuition motivating the method is that deep residual networks can efficiently capture high-dimensional function spaces and that random quadrature sampling aligns naturally with SGD, offering adaptivity and scalability. Empirical results demonstrate applications to Poisson equations in both low and high dimensions, Neumann boundary conditions, and eigenvalue problems \citep{evans2022partial}. The method achieved errors comparable to or smaller than finite difference baselines \citep{thomas1995finite} while using fewer parameters, though challenges such as boundary condition handling and degradation of accuracy in higher dimensions were noted.

The paper in \citep{sirignano2018dgm} introduces the Deep Galerkin Method (DGM) for solving high-dimensional PDEs. In this approach, a deep neural network parameterizes the candidate solution, and the PDE residual together with initial and boundary conditions are enforced directly in the loss function by sampling points from the domain. Because the network architecture is generic and the problem is specified entirely through the loss without requiring training data, the method belongs to the architecture-agnostic dNN category. The intuition motivating the method is that stochastic sampling of points provides an efficient mesh-free alternative to classical grid methods, avoiding the curse of dimensionality, while deep networks offer flexible approximation capacity for smooth PDE solutions. Empirical results on a variety of PDEs, including Black–Scholes, Hamilton–Jacobi–Bellman, and nonlinear reaction-diffusion equations \citep{evans2022partial}, show that the method produces solutions in high dimensions where finite difference methods are impractical. The results also indicate that stability and accuracy depend on careful sampling strategies and optimization, highlighting both the promise and the limitations of the approach.



\section{dataless NN-inspired methods}

In this section, we provide an overview of training-data-free methods that do not use a neural network explicitly but are inspired by neural networks tools that were initially developed and/or adopted for neural networks such as the use of activation function (e.g., Gumbel softmax \citep{jang2017categorical} and softplus \citep{goodfellow2016deep}) and the use of GPU-based parallelization. Similar to the previous section, each subsection corresponds to an application or problems. Since there is not a specific neural networks here, we will not use the dNN categorizes from Section~\ref{sec: def and cat}. 

\subsection{NP-hard graph combinatorial problems}


The paper in \citep{alkhouriicml} introduced a clique-informed differentiable quadratic formulation for MIS problem that augments the standard MIS quadratic objective with a complementary-graph (maximum-clique) term, optimized over box constraints using projected momentum-based gradient descent with parallel initializations. It also introduces a single-step projected-gradient criterion to check maximality efficiently, which the authors use to accelerate the implementation. The intuition is that the clique term both discourages picking adjacent vertices and counters overly sparse solutions, while gradient information plus GPU parallelism provides a practical exploration. On the analysis side, the paper derives conditions on the edge-penalty and clique-term parameters under which every maximal independent set is a local minimizer; shows all local minimizers are binary and correspond to maximal sets; and establishes that any non-binary stationary point is a saddle. Empirically, across multiple benchmarks, the method attains competitive or larger MIS sizes than recent learning, sampling, and exact/heuristic baselines under comparable or shorter run-time budgets. The authors note sensitivity to hyper-parameters and that sparse graphs still favor specialized heuristics such as ReduMIS \citep{lamm2016redumis}. \textcolor{black}{The idea of using GPU parallel processing with relaxed quadratic formulations was later extended to the Maximum Cut problem in \cite{alkhouri2025scalable} where similar empirical observations are noted. }

The paper in \citep{sun2023revisiting} proposes iSCO, a training-data-free sampler for combinatorial optimization that frames each problem as an energy-based objective and simulates discrete Langevin dynamics \citep{tome2015stochastic} using GPUs; key to the method is estimating local objective/probability ratios with gradients and evaluating large neighborhoods in parallel on accelerators. It generalizes the Path Auxiliary Sampler in \citep{sun2021path} to categorical variables so multiple coordinates can be updated at once and then corrected with a Metropolis–Hastings step, which accelerates otherwise slow Gibbs-style moves. Empirically, across graph-partitioning, MaxCut, routing tasks, and MIS, iSCO often achieved favorable speed–quality trade-offs relative to recent training-data-intensive solvers. Ablations show sensitivity to the initial temperature and annealing schedule, and diminishing returns from adding more chains. 

This GPU-based sampling method (i.e., iSCO) was later extended in \citep{lireheated} where the authors introduced a mechanism to mitigate the “wandering in contours” phenomena.

\subsection{Scheduling}

The paper in \citep{liu2024differentiable} proposed a differentiable combinatorial scheduling framework that casts latency-constrained, resource-aware scheduling as stochastic optimization over one-hot stage assignments, using Gumbel-Softmax activation function to relax the discrete variables and a constrained Gumbel trick to encode system-of-difference-constraint (SDC) inequalities in a differentiable way. The search space is vectorized per node under a latency bound, feasibility is enforced by a cumulative-sum–based transformation of one-hot vectors, and the objective couples an entropy-style surrogate for peak memory with an inter-stage communication term so the schedule can be optimized with gradient descent on GPUs. The intuition is that differentiable sampling offers a highly parallel, mesh-free alternative to constraints programming and integer LP. Empirically, on synthetic and real-work benchmarks, the proposed method shows optimization efficiency in many cases, surpassing CPLEX \citep{cplex2017}, Gurobi \citep{gurobi2023}, and the most recent CP-SAT \citep{cpsatlp} method under a time budget. 

More recently, the work in \citep{bara2025dataless} introduced a differentiable method for the resource-constrained project scheduling problem \citep{BRUCKER19993}. The method proposed to relax binary variables using the softplus activation function. Then, relu-based terms are used for the two problem-specific penalties. The paper provides preliminary empirical results showing that memory usage is minimal.   

\section{\textcolor{black}{Open questions and future directions}}

Despite the growing body of work on dataless neural networks, many theoretical and practical aspects remain open. From a theoretical perspective, a key challenge is to characterize the optimization landscapes induced by dNN parameterizations—specifically, understanding when the neural embedding of a problem instance guarantees convergence to globally or locally optimal solutions. Related to this is the need for a formal connection between dNN parameter spaces and classical convex or lifted formulations, which could provide insights into expressivity, stability, and implicit regularization.

On the algorithmic side, an important direction is improving scalability and efficiency for large or high-dimensional problem instances.

\bibliography{refs.bib}

\begin{thebibliography}{56}
\providecommand{\natexlab}[1]{#1}

\bibitem[{Adam et~al.(2014)}]{adam2014method}
Adam, K. D. B.~J.; et~al. 2014.
\newblock A method for stochastic optimization.
\newblock \emph{arXiv preprint arXiv:1412.6980}, 1412(6).

\bibitem[{Alkhouri et~al.(2025{\natexlab{a}})Alkhouri, Bell, Ghosh, Liang, Wang, and Ravishankar}]{alkhouri2025understanding}
Alkhouri, I.; Bell, E.; Ghosh, A.; Liang, S.; Wang, R.; and Ravishankar, S. 2025{\natexlab{a}}.
\newblock Understanding Untrained Deep Models for Inverse Problems: Algorithms and Theory.
\newblock \emph{IEEE Signal Processing Magazine}.

\bibitem[{Alkhouri et~al.(2024{\natexlab{a}})Alkhouri, Denmat, Li, Yu, Liu, Wang, and Velasquez}]{alkhouri2024dataless}
Alkhouri, I.; Denmat, C.~L.; Li, Y.; Yu, C.; Liu, J.; Wang, R.; and Velasquez, A. 2024{\natexlab{a}}.
\newblock Dataless Quadratic Neural Networks for the Maximum Independent Set Problem.
\newblock \emph{arXiv preprint arXiv:2406.19532}.

\bibitem[{Alkhouri et~al.(2025{\natexlab{b}})Alkhouri, Le~Denmat, Li, Yu, Liu, Wang, and Velasquez}]{alkhouriicml}
Alkhouri, I.; Le~Denmat, C.; Li, Y.; Yu, C.; Liu, J.; Wang, R.; and Velasquez, A. 2025{\natexlab{b}}.
\newblock Differentiable Quadratic Optimization For the Maximum Independent Set Problem.
\newblock In \emph{Forty-second International Conference on Machine Learning}.

\bibitem[{Alkhouri et~al.(2024{\natexlab{b}})Alkhouri, Liang, Bell, Qu, Wang, and Ravishankar}]{alkhouri2024image}
Alkhouri, I.; Liang, S.; Bell, E.; Qu, Q.; Wang, R.; and Ravishankar, S. 2024{\natexlab{b}}.
\newblock Image Reconstruction Via Autoencoding Sequential Deep Image Prior.
\newblock \emph{Advances in Neural Information Processing Systems}, 37: 18988--19012.

\bibitem[{Alkhouri et~al.(2025{\natexlab{c}})Alkhouri, Wu, Yu, Liu, Wang, and Velasquez}]{alkhouri2025scalable}
Alkhouri, I.; Wu, M.; Yu, C.; Liu, J.; Wang, R.; and Velasquez, A. 2025{\natexlab{c}}.
\newblock A Scalable Lift-and-Project Differentiable Approach For the Maximum Cut Problem.
\newblock \emph{arXiv preprint arXiv:2509.18612}.

\bibitem[{Alkhouri, Atia, and Velasquez(2022)}]{alkhouriRelu}
Alkhouri, I.~R.; Atia, G.~K.; and Velasquez, A. 2022.
\newblock A differentiable approach to the maximum independent set problem using dataless neural networks.
\newblock \emph{Neural Networks}, 155: 168--176.

\bibitem[{Alkhouri, Velasquez, and Atia(2022)}]{alkhouri2022boss}
Alkhouri, I.~R.; Velasquez, A.; and Atia, G.~K. 2022.
\newblock Boss: Bidirectional one-shot synthesis of adversarial examples.
\newblock In \emph{2022 IEEE 32nd International Workshop on Machine Learning for Signal Processing (MLSP)}, 1--6. IEEE.

\bibitem[{Bara(2025)}]{bara2025dataless}
Bara, M. 2025.
\newblock Dataless Neural Networks for Resource-Constrained Project Scheduling.
\newblock \emph{arXiv preprint arXiv:2507.05322}.

\bibitem[{Ben-Tal and Nemirovski(2001)}]{ben2001lectures}
Ben-Tal, A.; and Nemirovski, A. 2001.
\newblock \emph{Lectures on Modern Convex Optimization: Analysis, Algorithms, and Engineering Applications}.
\newblock Philadelphia, PA: SIAM.

\bibitem[{Brown et~al.(2020)Brown, Mann, Ryder, Subbiah, Kaplan, Dhariwal, Neelakantan, Shyam, Sastry, Askell, Agarwal, Herbert-Voss, Krueger, Henighan, Child, Ramesh, Ziegler, Wu, Winter, Hesse, Chen, Sigler, Litwin, Gray, Chess, Clark, Berner, McCandlish, Radford, Sutskever, and Amodei}]{brown2020language}
Brown, T.~B.; Mann, B.; Ryder, N.; Subbiah, M.; Kaplan, J.; Dhariwal, P.; Neelakantan, A.; Shyam, P.; Sastry, G.; Askell, A.; Agarwal, S.; Herbert-Voss, A.; Krueger, G.; Henighan, T.; Child, R.; Ramesh, A.; Ziegler, D.~M.; Wu, J.; Winter, C.; Hesse, C.; Chen, M.; Sigler, E.; Litwin, M.; Gray, S.; Chess, B.; Clark, J.; Berner, C.; McCandlish, S.; Radford, A.; Sutskever, I.; and Amodei, D. 2020.
\newblock Language models are few-shot learners.
\newblock In \emph{Advances in Neural Information Processing Systems}, volume~33, 1877--1901.

\bibitem[{Brucker et~al.(1999)Brucker, Drexl, Möhring, Neumann, and Pesch}]{BRUCKER19993}
Brucker, P.; Drexl, A.; Möhring, R.; Neumann, K.; and Pesch, E. 1999.
\newblock Resource-constrained project scheduling: Notation, classification, models, and methods.
\newblock \emph{European Journal of Operational Research}, 112(1): 3--41.

\bibitem[{Cobb et~al.(2024)Cobb, Matejek, Elenius, Roy, and Jha}]{cobb2024direct}
Cobb, A.~D.; Matejek, B.; Elenius, D.; Roy, A.; and Jha, S. 2024.
\newblock Direct amortized likelihood ratio estimation.
\newblock In \emph{Proceedings of the AAAI Conference on Artificial Intelligence}, volume~38, 20362--20369.

\bibitem[{Dantzig(1951)}]{dantzig1951maximization}
Dantzig, G.~B. 1951.
\newblock Maximization of a linear function of variables subject to linear inequalities.
\newblock \emph{Activity analysis of production and allocation}, 13: 339--347.

\bibitem[{Evans(2022)}]{evans2022partial}
Evans, L.~C. 2022.
\newblock \emph{Partial differential equations}, volume~19.
\newblock American mathematical society.

\bibitem[{Gilmer et~al.(2017)Gilmer, Schoenholz, Riley, Vinyals, and Dahl}]{gilmer2017neural}
Gilmer, J.; Schoenholz, S.~S.; Riley, P.~F.; Vinyals, O.; and Dahl, G.~E. 2017.
\newblock Neural message passing for quantum chemistry.
\newblock In \emph{International conference on machine learning}, 1263--1272. Pmlr.

\bibitem[{Goodfellow, Bengio, and Courville(2016)}]{goodfellow2016deep}
Goodfellow, I.; Bengio, Y.; and Courville, A. 2016.
\newblock \emph{Deep Learning}.
\newblock MIT Press.

\bibitem[{{Google}()}]{cpsatlp}
{Google}. ????
\newblock \emph{CP-SAT}.

\bibitem[{{Gurobi Optimization, LLC}()}]{gurobi2023}
{Gurobi Optimization, LLC}. ????
\newblock \emph{Gurobi Optimizer Reference Manual}.

\bibitem[{Ho, Jain, and Abbeel(2020)}]{ho2020denoising}
Ho, J.; Jain, A.; and Abbeel, P. 2020.
\newblock Denoising Diffusion Probabilistic Models.
\newblock In \emph{Advances in Neural Information Processing Systems}, volume~33, 6840--6851.

\bibitem[{Hosny and Reda(2024)}]{hosny2024torchmsat}
Hosny, A.; and Reda, S. 2024.
\newblock torchmSAT: A GPU-Accelerated Approximation To The Maximum Satisfiability Problem.
\newblock \emph{arXiv preprint arXiv:2402.03640}.

\bibitem[{{IBM}()}]{cplex2017}
{IBM}. ????
\newblock \emph{IBM ILOG CPLEX Optimization Studio CPLEX User's Manual}.
\newblock IBM, Armonk, NY.

\bibitem[{Ichikawa(2024)}]{ichikawa2024controlling}
Ichikawa, Y. 2024.
\newblock Controlling continuous relaxation for combinatorial optimization.
\newblock \emph{Advances in Neural Information Processing Systems}, 37: 47189--47216.

\bibitem[{Jang, Gu, and Poole(2017)}]{jang2017categorical}
Jang, E.; Gu, S.; and Poole, B. 2017.
\newblock Categorical Reparameterization with Gumbel-Softmax.
\newblock In \emph{International Conference on Learning Representations (ICLR)}.

\bibitem[{Jena, Subramani, and Velasquez(2024)}]{relu_dnn_mani_several}
Jena, S.~K.; Subramani, K.; and Velasquez, A. 2024.
\newblock A Differential Approach for Several NP-hard Optimization Problems.
\newblock In Barneva, R.~P.; Brimkov, V.~E.; Gentile, C.; and Pacchiano, A., eds., \emph{Artificial Intelligence and Image Analysis}, 68--80. Cham: Springer Nature Switzerland.
\newblock ISBN 978-3-031-63735-3.

\bibitem[{Jumper et~al.(2021)Jumper, Evans, Pritzel, Green, Figurnov, Ronneberger, Tunyasuvunakool, Bates, {\v Z}{\'\i}dek, Potapenko, Bridgland, Meyer, Kohl, Ballard, Cowie, Romera-Paredes, Nikolov, Jain, Adler, Back, Petersen, Reiman, Clancy, Zielinski, Steinegger, Pacholska, Berghammer, Bodenstein, Silver, Vinyals, Senior, Kavukcuoglu, Kohli, and Hassabis}]{jumper2021highly}
Jumper, J.; Evans, R.; Pritzel, A.; Green, T.; Figurnov, M.; Ronneberger, O.; Tunyasuvunakool, K.; Bates, R.; {\v Z}{\'\i}dek, A.; Potapenko, A.; Bridgland, A.; Meyer, C.; Kohl, S. A.~A.; Ballard, A.~J.; Cowie, A.; Romera-Paredes, B.; Nikolov, S.; Jain, R.; Adler, J.; Back, T.; Petersen, S.; Reiman, D.; Clancy, E.; Zielinski, M.; Steinegger, M.; Pacholska, M.; Berghammer, T.; Bodenstein, S.; Silver, D.; Vinyals, O.; Senior, A.~W.; Kavukcuoglu, K.; Kohli, P.; and Hassabis, D. 2021.
\newblock Highly accurate protein structure prediction with AlphaFold.
\newblock \emph{Nature}, 596(7873): 583--589.

\bibitem[{Kim and Fridovich-Keil(2025)}]{kim2025grids}
Kim, N.; and Fridovich-Keil, S. 2025.
\newblock Grids Often Outperform Implicit Neural Representations.
\newblock \emph{arXiv preprint arXiv:2506.11139}.

\bibitem[{Kunc and Kl{\'e}ma(2024)}]{kunc2024three}
Kunc, V.; and Kl{\'e}ma, J. 2024.
\newblock Three decades of activations: A comprehensive survey of 400 activation functions for neural networks.
\newblock \emph{arXiv preprint arXiv:2402.09092}.

\bibitem[{Lamm et~al.(2016)Lamm, Sanders, Schulz, Strash, and Werneck}]{lamm2016redumis}
Lamm, S.; Sanders, P.; Schulz, C.; Strash, D.; and Werneck, R.~F. 2016.
\newblock Accelerating Local Search for the Maximum Independent Set Problem.
\newblock In \emph{Proceedings of the 2016 ALENEX (Algorithm Engineering and Experiments) Symposium (SEA 2016)}.

\bibitem[{Li and Zhang(2025)}]{lireheated}
Li, M.; and Zhang, R. 2025.
\newblock Reheated Gradient-based Discrete Sampling for Combinatorial Optimization.
\newblock \emph{Transactions on Machine Learning Research}.

\bibitem[{Liang et~al.(2025{\natexlab{a}})Liang, Alkhouri, Gautam, Qu, and Ravishankar}]{liang2025ugodit}
Liang, S.; Alkhouri, I.~R.; Gautam, S.; Qu, Q.; and Ravishankar, S. 2025{\natexlab{a}}.
\newblock UGoDIT: Unsupervised Group Deep Image Prior Via Transferable Weights.
\newblock \emph{arXiv preprint arXiv:2505.11720}.

\bibitem[{Liang et~al.(2025{\natexlab{b}})Liang, Bell, Qu, Wang, and Ravishankar}]{shijun_self_guided}
Liang, S.; Bell, E.; Qu, Q.; Wang, R.; and Ravishankar, S. 2025{\natexlab{b}}.
\newblock Analysis of Deep Image Prior and Exploiting Self-Guidance for Image Reconstruction.
\newblock \emph{IEEE Transactions on Computational Imaging}, 11: 435--451.

\bibitem[{Liu et~al.(2024)Liu, Li, Yin, Zhang, and Yu}]{liu2024differentiable}
Liu, M.; Li, Y.; Yin, J.; Zhang, Z.; and Yu, C. 2024.
\newblock Differentiable Combinatorial Scheduling at Scale.
\newblock \emph{arXiv preprint arXiv:2406.06593}.

\bibitem[{Nielsen(2015)}]{nielsen2015neural}
Nielsen, M.~A. 2015.
\newblock \emph{Neural networks and deep learning}, volume~25.
\newblock Determination press San Francisco, CA, USA.

\bibitem[{Novikov et~al.(2025)Novikov, Vũ, Eisenberger, Dupont, Huang, Wagner, Shirobokov, Kozlovskii, Ruiz, Mehrabian, Kumar, See, Chaudhuri, Holland, Davies, Nowozin, Kohli, and Balog}]{novikov2025alphaevolve}
Novikov, A.; Vũ, N.; Eisenberger, M.; Dupont, E.; Huang, P.; Wagner, A.~Z.; Shirobokov, S.; Kozlovskii, B.; Ruiz, F. J.~R.; Mehrabian, A.; Kumar, M.~P.; See, A.; Chaudhuri, S.; Holland, G.; Davies, A.; Nowozin, S.; Kohli, P.; and Balog, M. 2025.
\newblock AlphaEvolve: A coding agent for scientific and algorithmic discovery.
\newblock Tech.~report, Google DeepMind.

\bibitem[{Oppenheim, Willsky, and Nawab(1997)}]{oppenheim1997signals}
Oppenheim, A.~V.; Willsky, A.~S.; and Nawab, S.~H. 1997.
\newblock \emph{Signals \& systems}.
\newblock Pearson Educaci{\'o}n.

\bibitem[{Pourpanah et~al.(2022)Pourpanah, Abdar, Luo, Zhou, Wang, Lim, Wang, and Wu}]{pourpanah2022review}
Pourpanah, F.; Abdar, M.; Luo, Y.; Zhou, X.; Wang, R.; Lim, C.~P.; Wang, X.-Z.; and Wu, Q.~J. 2022.
\newblock A review of generalized zero-shot learning methods.
\newblock \emph{IEEE transactions on pattern analysis and machine intelligence}, 45(4): 4051--4070.

\bibitem[{Radford et~al.(2018)Radford, Narasimhan, Salimans, and Sutskever}]{radford2018improving}
Radford, A.; Narasimhan, K.; Salimans, T.; and Sutskever, I. 2018.
\newblock Improving Language Understanding by Generative Pre-Training.
\newblock \url{https://cdn.openai.com/research-covers/language-unsupervised/language_understanding_paper.pdf}.
\newblock OpenAI Technical Report.

\bibitem[{Radford et~al.(2019)Radford, Wu, Child, Luan, Amodei, and Sutskever}]{radford2019language}
Radford, A.; Wu, J.; Child, R.; Luan, D.; Amodei, D.; and Sutskever, I. 2019.
\newblock Language Models are Unsupervised Multitask Learners.
\newblock \url{https://cdn.openai.com/better-language-models/language_models_are_unsupervised_multitask_learners.pdf}.
\newblock OpenAI Technical Report.

\bibitem[{Rombach et~al.(2022)Rombach, Blattmann, Lorenz, Esser, and Ommer}]{rombach2022high}
Rombach, R.; Blattmann, A.; Lorenz, D.; Esser, P.; and Ommer, B. 2022.
\newblock High-resolution image synthesis with latent diffusion models.
\newblock In \emph{Proceedings of the IEEE/CVF Conference on Computer Vision and Pattern Recognition (CVPR)}, 10684--10695.

\bibitem[{Rumelhart, Hinton, and Williams(1986)}]{rumelhart1986learning}
Rumelhart, D.~E.; Hinton, G.~E.; and Williams, R.~J. 1986.
\newblock Learning representations by back-propagating errors.
\newblock \emph{nature}, 323(6088): 533--536.

\bibitem[{Schuetz, Brubaker, and Katzgraber(2022)}]{schuetz2022combinatorial}
Schuetz, M.~J.; Brubaker, J.~K.; and Katzgraber, H.~G. 2022.
\newblock Combinatorial optimization with physics-inspired graph neural networks.
\newblock \emph{Nature Machine Intelligence}, 4(4): 367--377.

\bibitem[{Sirignano and Spiliopoulos(2018)}]{sirignano2018dgm}
Sirignano, J.; and Spiliopoulos, K. 2018.
\newblock DGM: A deep learning algorithm for solving partial differential equations.
\newblock \emph{Journal of computational physics}, 375: 1339--1364.

\bibitem[{Sitzmann et~al.(2020)Sitzmann, Martel, Bergman, Lindell, and Wetzstein}]{sitzmann2020implicit}
Sitzmann, V.; Martel, J.; Bergman, A.; Lindell, D.; and Wetzstein, G. 2020.
\newblock Implicit neural representations with periodic activation functions.
\newblock \emph{Advances in neural information processing systems}, 33: 7462--7473.

\bibitem[{Song and Ermon(2019)}]{song2019generative}
Song, Y.; and Ermon, S. 2019.
\newblock Generative Modeling by Estimating Gradients of the Data Distribution.
\newblock In \emph{Advances in Neural Information Processing Systems}, volume~32.

\bibitem[{Song et~al.(2023)Song, Wang, Cai, Mondal, and Sahoo}]{song2023comprehensive}
Song, Y.; Wang, T.; Cai, P.; Mondal, S.~K.; and Sahoo, J.~P. 2023.
\newblock A comprehensive survey of few-shot learning: Evolution, applications, challenges, and opportunities.
\newblock \emph{ACM Computing Surveys}, 55(13s): 1--40.

\bibitem[{Sun et~al.(2021)Sun, Dai, Xia, and Ramamurthy}]{sun2021path}
Sun, H.; Dai, H.; Xia, W.; and Ramamurthy, A. 2021.
\newblock Path auxiliary proposal for MCMC in discrete space.
\newblock In \emph{International Conference on Learning Representations}.

\bibitem[{Sun et~al.(2023)Sun, Goshvadi, Nova, Schuurmans, and Dai}]{sun2023revisiting}
Sun, H.; Goshvadi, K.; Nova, A.; Schuurmans, D.; and Dai, H. 2023.
\newblock Revisiting sampling for combinatorial optimization.
\newblock In \emph{International Conference on Machine Learning}, 32859--32874. PMLR.

\bibitem[{Thomas(1995)}]{thomas1995finite}
Thomas, J.~W. 1995.
\newblock \emph{Numerical Partial Differential Equations: Finite Difference Methods}, volume~22 of \emph{Texts in Applied Mathematics}.
\newblock Springer.

\bibitem[{Tom{\'{e}} and de~Oliveira(2015)}]{tome2015stochastic}
Tom{\'{e}}, T.; and de~Oliveira, M.~J. 2015.
\newblock \emph{Stochastic Dynamics and Irreversibility}.
\newblock Graduate Texts in Physics. Springer Cham.

\bibitem[{Ulyanov, Vedaldi, and Lempitsky(2018)}]{ulyanov2018deep}
Ulyanov, D.; Vedaldi, A.; and Lempitsky, V. 2018.
\newblock Deep image prior.
\newblock In \emph{Proceedings of the IEEE conference on computer vision and pattern recognition}, 9446--9454.

\bibitem[{Wu and Tam(1999)}]{798504}
Wu, A.; and Tam, P. 1999.
\newblock Using neural network method computes quadratic optimization problems.
\newblock In \emph{Proceedings Third International Conference on Computational Intelligence and Multimedia Applications. ICCIMA'99 (Cat. No.PR00300)}, 70--74.

\bibitem[{Yaman, Hosseini, and Akcakaya()}]{yamanzero}
Yaman, B.; Hosseini, S. A.~H.; and Akcakaya, M. ????
\newblock Zero-Shot Self-Supervised Learning for MRI Reconstruction.
\newblock In \emph{International Conference on Learning Representations}.

\bibitem[{Yu et~al.(2018)}]{yu2018deep}
Yu, B.; et~al. 2018.
\newblock The deep Ritz method: a deep learning-based numerical algorithm for solving variational problems.
\newblock \emph{Communications in Mathematics and Statistics}, 6(1): 1--12.

\bibitem[{Zak, Upatising, and Hui(1995)}]{NNforLPs1995}
Zak, S.; Upatising, V.; and Hui, S. 1995.
\newblock Solving linear programming problems with neural networks: a comparative study.
\newblock \emph{IEEE Transactions on Neural Networks}, 6(1): 94--104.

\bibitem[{Zhang et~al.(2024)Zhang, Zhen, Yuan, and Yu}]{zhang2024diffsat}
Zhang, Y.; Zhen, H.-L.; Yuan, M.; and Yu, B. 2024.
\newblock DiffSAT: Differential MaxSAT Layer for SAT Solving.
\newblock In \emph{Proceedings of the 43rd IEEE/ACM International Conference on Computer-Aided Design}, 1--7.

\end{thebibliography}


\end{document}